\documentclass[sn-basic,Numbered,iicol]{sn-jnl}

\usepackage{amsmath}
\usepackage{amssymb}
\usepackage{booktabs}
\usepackage{bm}          
\usepackage{microtype}

\theoremstyle{thmstyleone}

\theoremstyle{thmstyletwo}

\theoremstyle{thmstylethree}

\makeatletter
\def\sectionfont{\reset@font\fontfamily{\rmdefault}%
  \fontsize{11.5bp}{13.5bp}\bfseries\selectfont\raggedright\boldmath}
\def\subsectionfont{\reset@font\fontfamily{\rmdefault}%
  \fontsize{10.5bp}{12.5bp}\bfseries\selectfont\raggedright\boldmath}
\def\subsubsectionfont{\reset@font%
  \fontsize{10bp}{12bp}\bfseries\selectfont\raggedright\boldmath}
\makeatother

\begin{document}

\title[Fuzzy Control with AIRL]{Integrating AIRL-Based Reward Estimation into a Control Method Using Fuzzy Inference and Policy Gradient}

\author*[1]{\fnm{Seiji}\sur{Ishihara}}\email{ishihara\_s@mail.dendai.ac.jp}

\affil*[1]{\orgdiv{School of Science and Engineering},
           \orgname{Tokyo Denki University},
           \orgaddress{\city{Saitama}, \country{Japan}}}

\abstract{A method integrating policy gradient with fuzzy inference has been proposed. 
This method, hereafter referred to as the base model, has high flexibility with respect to the configuration of fuzzy inference rules and fuzzy representation.
 It can also automatically adjust the rule weights and the membership functions in the antecedent and consequent parts using the same framework for learning.
In methods that integrate reinforcement learning with fuzzy inference, however, the reward function must be designed in advance.
It is therefore useful to add learning-based reward estimation to the base model.
 This paper proposes a fuzzy control method that integrates a reward estimation mechanism based on the adversarial inverse reinforcement learning (AIRL) framework into the base model.
The generative adversarial nets (GAN) framework adopted in AIRL, however, tends to suffer from unstable learning. 
To address this, the proposed method treats the policy of the base model as the generator and also uses the set of membership functions in the policy as the cost function of the discriminator.
This configuration allows the discriminator update for reward estimation and the generator update for policy estimation to be performed simultaneously on the parameters shared by both, without estimating a value function. 
In simulations of an illustrative vehicle speed control task, the proposed method acquired policies with high generalization performance through stable and efficient learning.}


\maketitle

\section*{Article Highlights}
\begin{itemize}
\item Fuzzy rule parameters that both decide an action and evaluate it are acquired automatically through learning.
\item Sharing the parameters lets both kinds of learning proceed at once, without another function that evaluates an action.
\item Desirable speed control was learned stably with far less training than by a conventional alternating learning method.
\end{itemize}

\section{Introduction}\label{sec:intro}

Control based on fuzzy inference has high explainability in its process because the knowledge representation is readily understandable by humans.
Methods for integrating reinforcement learning into such fuzzy control systems have been studied \cite{Glorennec,Horiuchi,Jouffe,Er,Wang,Aghaei,Imanina,IshiharaMF2024}.
These methods automatically adjust the parameters constituting if-then rules over continuous state and action spaces, using rewards designed according to the control task.

 Among these is a method \cite{Imanina,IshiharaMF2024}, hereafter referred to as the base model, using the policy gradient method \cite{Williams,Baxter,IshiharaPursuit} to learn the weights representing the reliability of each fuzzy inference rule (rule weights) and the parameters that specify the shape of each membership function in the antecedent and consequent parts.
 This method has high flexibility with respect to rule configuration and fuzzy representation. 
It can also use the same learning framework for the automatic adjustment of the rule weights and that of the membership functions.
In methods that integrate reinforcement learning with fuzzy inference, however, the reward function must be designed in advance.

 Meanwhile, studies on imitation learning and inverse reinforcement learning (IRL) have been reported in recent years \cite{Arora,Zare}.
 These methods acquire a policy equivalent to that of an expert, or an effective reward function, from the state-action trajectories generated by the expert's strategy for a given task.
The approach of formulating the task as an optimization problem based on the maximum entropy principle and solving it by a gradient method \cite{Ziebart} can be readily combined with neural networks.
Among methods based on this approach, adversarial IRL (AIRL) \cite{Fu} is efficient in that it updates the policy and the reward alternately and partially, and in that it takes as input the state-action pair at each time step. 
It also has the advantage of extracting a reward function that does not include environment dynamics.
The generative adversarial nets (GAN) framework \cite{Goodfellow} adopted in AIRL, however, tends to suffer from unstable learning.

In such an integration of reinforcement learning and fuzzy inference, it is useful to add a mechanism that estimates the reward function through learning.
This paper therefore proposes a fuzzy control method that integrates a reward estimation mechanism based on the AIRL framework into the base model.

The proposed method does not adopt the AIRL discriminator as it stands.
Instead, to make the learning of both the reward function and the policy stable and efficient, it treats the policy of the base model as the generator and also uses the set of membership functions in the policy as the internal function of the discriminator, namely its cost function\footnote{ Since the internal function is not explicitly named in AIRL, this paper borrows the term from GCL \cite{Finn} and defines the cost function so that a larger value is more desirable.}.
This cost function approximates a quantity corresponding to the advantage of an action with respect to the optimal policy, using the set of membership functions rather than a value function.
In the learning stage, this allows the discriminator update for reward estimation and the generator update for policy estimation to be performed simultaneously on the parameters shared by both.
Since the discriminator and the generator share the parameters of the membership functions and no value function has to be estimated, learning becomes stable and efficient. 
The proposed method thus forms an integrated structure, in which the membership functions of the policy-gradient-based generator are also applied to the configuration of the discriminator.

Simulations of an illustrative vehicle speed control task show that the proposed method achieves stable and efficient learning and acquires policies with high generalization performance.

The main contributions of this paper are as follows. 
\begin{enumerate}
\item A fuzzy control method is proposed that integrates a reward estimation mechanism based on the AIRL framework into the base model. The reward function no longer has to be designed in advance.
\item A cost function of the discriminator is defined using the set of membership functions in the policy of the generator. The parameters of the membership functions are thus shared between the generator and the discriminator. Under the optimal policy, the cost function approximates a quantity corresponding to the advantage of an action without a value function.
\item The discriminator update for reward estimation and the generator update for policy estimation are performed simultaneously on the shared parameters, and no value function has to be estimated. Learning thus becomes stable and efficient.
\end{enumerate}

The paper is organized as follows.
Section~\ref{sec:relate} reviews related work on methods integrating reinforcement learning with fuzzy inference and on inverse reinforcement learning.
Section~\ref{sec:base-mdl} describes the base model.
Section~\ref{sec:AIRL} outlines AIRL.
Section~\ref{sec:pro-meth} presents the configuration of the generator and the discriminator in the proposed method, together with their learning rules.
Section~\ref{sec:task} explains the control task and its application.
Section~\ref{sec:exp} details the computer experiments and their results, and discusses the findings.
Section~\ref{sec:conc} concludes the paper.

\section{Related Work}
\label{sec:relate}

\subsection{Methods Integrating Reinforcement Learning with Fuzzy Inference}
\label{subsec:RL-fuz}

These integration methods can be broadly divided into two types according to the reinforcement learning they adopt.

Under the assumption of a Markov decision process (MDP) for the environment and the agent's policy, the first type uses value-based reinforcement learning.
Glorennec's fuzzy Q-learning \cite{Glorennec}, which combines Q-learning with fuzzy inference, and its extensions \cite{Horiuchi,Jouffe} belong to this type.
Methods of this type generally apply fuzzy representation to the state input variables of the antecedent part, while the parameter values or functions corresponding to the Q-value of each action are given as the output of the consequent part without fuzzy representation.
They also have no mechanism for adaptively strengthening or suppressing the rules themselves.
One method adds rules as needed according to the matching degree of the antecedent part or the temporal difference (TD) error during reinforcement learning \cite{Er}, but it does not weight the rules directly.

The second type uses policy-based reinforcement learning, specifically the policy gradient method \cite{Williams,Baxter,IshiharaPursuit}, which learns the policy directly by a gradient method without using a value function.
As in the first type, in one method \cite{Wang} that uses GPOMDP \cite{Baxter}, each rule consists of an antecedent part with fuzzy representation and a consequent part that outputs the parameters corresponding to the policy without fuzzy representation.
Another method \cite{Aghaei} has a consequent part that outputs the matching degree of a fuzzy representation defined by the parameters corresponding to the policy.
This method integrates policy gradient methods such as GPOMDP with fuzzy inference.
In these methods, however, the membership functions are fixed and the rule weights are not taken into account.
The base model \cite{Imanina,IshiharaMF2024}, by contrast, has high flexibility with respect to the configuration of fuzzy inference rules: it supports fuzzy representation not only in the antecedent part but also in the consequent part, and it allows rule weights corresponding to the reliability of each rule to be set.
It can also be applied as a general-purpose method that uses the same framework for learning to automatically acquire the reliability of each rule and the concept of fuzzy representation based on the shape of each membership function \cite{IshiharaMF2024}.

\subsection{Inverse Reinforcement Learning}
\label{subsec:IRL}

Inverse reinforcement learning is a type of imitation learning that automatically acquires an expert's policy from the expert's trajectories, and it differs from other methods of imitation learning in that it can estimate the reward function explicitly.
Multiple types of IRL methods have been studied \cite{Arora}.
Among them, methods combining a gradient method with maximum entropy IRL (ME-IRL) \cite{Ziebart}, a later approach, have been developed.
Finn et al. proposed guided cost learning (GCL) \cite{Finn}, which alternately updates the policy and the cost function and estimates the partition function through sampling.
GCL extends the applicability of ME-IRL and improves efficiency.
They subsequently showed that GCL can be reformulated within the GAN framework (GAN-GCL) \cite{Finn2}.
Fu et al. proposed AIRL \cite{Fu}, which takes as input the state-action pair at each time step rather than an entire trajectory and thereby learns more efficiently.
By incorporating a value function into the discriminator, AIRL also estimates a reward function that does not include environment dynamics.
In AIRL, the cost function of the discriminator has been shown to be equivalent to the advantage function under the optimal policy.
The GAN framework adopted in these methods, however, tends to suffer from unstable learning.
Approaches to stabilizing the training of GANs have therefore been explored \cite{Goodfellow2,Salimans,Karras}.
For adversarial learning in imitation learning and IRL, a method has also been proposed that constrains the information flow in the discriminator by an information bottleneck to maintain informative gradients \cite{Peng}.
In the context of fuzzy inference, no method has been reported that uses membership functions as the cost function to make learning in the AIRL framework stable and efficient.

\section{The Base Model}
\label{sec:base-mdl}

\subsection{Fuzzy Inference Rules}
\label{subsec:fuz-rule}

The following fuzzy inference rules are used to control the agent's behavior.
\begin{align}
&\text{Rule } i: \nonumber\\
&\quad \text{if } (x_1 \text{ is } A^i_1)
 \text{ and } \dots \text{ and } (x_M \text{ is } A^i_M) \nonumber\\
&\quad \text{then } (y_1 \text{ is } B^i_1)
 \text{ and } \dots \text{ and } (y_N \text{ is } B^i_N) \nonumber\\
&\quad \text{with } \eta_i
\label{eq:1}
\end{align}
Here, $i$ ($=1,2,\cdots,n_{\textrm{R}}$) is the rule number, $x_j$ ($j=1,2,\cdots,M$) is an input variable representing the state, and $y_k$ ($k=1,2,\cdots,N$) is an output variable representing the action.
$A^i_j$ and $B^i_k$ are linguistic fuzzy expressions for the variables $x_j$ and $y_k$ in the $j$-th antecedent part and the $k$-th consequent part of rule $i$, respectively.
$A^i_j(x_j;\alpha^i_j)$ and $B^i_k(y_k;\beta^i_k)$ are their membership functions, specified by the parameters $\alpha^i_j$ and $\beta^i_k$.
$\eta_i$ ($\geq 0$) is a parameter representing the rule weight.

\subsection{Objective Function and Policy}
\label{subsec:policy}

The agent decides its action at each time step so as to minimize the objective function
\begin{equation}
E(y;x)=-\sum_{i=1}^{n_{\textrm{R}}}\eta_iA^i(x;\alpha^i)B^i(y;\beta^i)
\label{eq:2}
\end{equation}
Here, the matching degrees of the input value $x$ in the antecedent part and of the output value $y$ in the consequent part of rule $i$ are defined as $A^i(x;\alpha^i)\equiv \prod_{j=1}^M A^i_j(x_j;\alpha^i_j)$ and $B^i(y;\beta^i)\equiv \prod_{k=1}^N B^i_k(y_k;\beta^i_k)$, respectively.
The objective function in Eq. (\ref{eq:2}) represents the total degree of support that the rules give to the output value $y$ for the input value $x$.

Deciding the output value $y(t)$ for the input value $x(t)$ at time $t$ is the policy of the agent, or learning system.
The policy is defined as a probability given by the Boltzmann distribution with temperature $T$, in which $E(y(t);x(t))$ is used as the energy function.
\begin{equation}
\pi(y(t)\!\mid\! x(t)) \equiv \frac{\exp(-E(y(t);x(t))/T)}{\sum_{y}\exp(-E(y;x(t))/T)}
\label{eq:3}
\end{equation}
The policy $\pi$ is thus stochastic, and $T$ represents the degree of randomness in deciding the action.

\subsection{Learning Rule for the Parameters by the Policy Gradient Method}
\label{subsec:LR-base-mdl}

The same reinforcement learning framework applies to all the parameters $\mu\in\{\alpha,\beta,\eta\}$.
The policy gradient method gives the following learning rule, which maximizes the expected reward $\mathbb{E}_{\tau\sim\pi}[r(\tau)]$ for a trajectory $\tau=(x(0),y(0),\cdots,x(L-1),y(L-1))$ in an episode.
\begin{equation}
\Delta\mu = \varepsilon \mathbb{E}_{\tau\sim\pi}\left[r(\tau)\sum_{t=0}^{L-1}e_{\mu}(t)\right]
\label{eq:4}
\end{equation}
Here, $L$ is the episode length and $\varepsilon$ ($>0$) is the learning coefficient.
The characteristic eligibility \cite{Williams} $e_{\mu}(t)$ is given by the following expressions \cite{IshiharaPursuit}.
\begin{align}
&e_{\mu}(t) \equiv \frac{\partial}{\partial \mu}\ln\pi\left( y(t)\!\mid\! x(t)\right)
\label{eq:5}\\
&= -\frac{1}{T}\!\left\{\!\frac{\partial E(y(t);x(t))}{\partial \mu}-\mathbb{E}_{\pi(y\mid x(t))}\!\!\left[\!\frac{\partial E(y;x(t))}{\partial \mu}\!\right]\!\right\}
\label{eq:6}
\end{align}
Here, $\mathbb{E}_{\pi(y\mid x)}[f(y)] \equiv \sum_{y}f(y)\ \pi(y\!\mid\! x)$.

\section{Adversarial Inverse Reinforcement Learning}
\label{sec:AIRL}

AIRL is an IRL method that adopts the GAN framework, which consists of a discriminator and a generator.
To obtain the optimal policy, AIRL alternately updates the discriminator and the policy in the generator.
Specifically, with the expert trajectories prepared in advance, the process repeats a cycle consisting of generating sample trajectories with the policy of the generator, updating the discriminator, estimating the reward from the updated discriminator, and updating the policy in the generator.

\subsection{Discriminator}
\label{subsec:disc}

Let $x'$ denote the state one step ahead, reached from state $x$ by action $y$.
The output of the discriminator defined using the cost function $f_{\theta,\phi}(x,y,x')$ is given by
\begin{equation}
D(x,y,x') = \frac{\exp(f_{\theta,\phi}(x,y,x'))}{\exp(f_{\theta,\phi}(x,y,x'))+\pi(y\!\mid\! x)}
\label{eq:7}
\end{equation}
The cost function is specified by the parameters $\theta$ and $\phi$ and is decomposed as follows:
\begin{equation}
f_{\theta,\phi}(x,y,x') = g(x,y;\theta)+\kappa h(x';\phi)-h(x;\phi)
\label{eq:8}
\end{equation}
Here, $g$ is an approximation of the reward function, $h$ is an approximation of the state value function, and $\kappa$ is the discount rate.
Under the optimal policy, $f_{\theta,\phi}(x,y,x')$ has been shown to be equivalent to the advantage function.

The value of $D(x,y,x')$ corresponds to the probability that the input $(x,y,x')$ is contained in the expert trajectory $\tau_{\textrm{D}}$, the demonstration.
It is also the probability that the input is not contained in the trajectory $\tau_{\textrm{S}}$ , the sample, generated by the actions that the generator outputs following the policy $\pi(y\!\mid\! x)$.
Let $\mathcal{D}_{t}$ denote the set of element triples $(x(t),y(t),x(t+1))$ based at time $t$ in the demonstration $\tau_{\textrm{D}}$.
For correct discrimination between the demonstration and the sample, the parameters of the discriminator are learned so as to minimize the following cross-entropy loss function.
\begin{align}
\mathcal{L}(\theta,\phi) = \sum_{t=0}^{L-1}&\left\{-\mathbb{E}_{\mathcal{D}_{t}}\!\left[\ln(D(x,y,x'))\right]\right.\nonumber\\
&\ \left.-\mathbb{E}_{\pi_t}\!\left[\ln(1-D(x,y,x'))\right]\right\}
\label{eq:9}
\end{align}
Here, $\pi_t$ denotes the distribution of $(x(t),y(t),x(t+1))$ generated by rolling out the policy $\pi$.
The gradient-based learning rule for the discriminator is given by
\begin{align}
\Delta\lambda = \delta\!\sum_{t=0}^{L-1}&\!\left\{\!\mathbb{E}_{\mathcal{D}_{t}}\!\!\left[\!(1-D(x,y,x'))\frac{\partial}{\partial\lambda}f_{\theta,\phi}(x,y,x')\!\right]\right.\nonumber\\
&\ \left.-\mathbb{E}_{\pi_t}\!\!\left[D(x,y,x')\frac{\partial}{\partial\lambda}f_{\theta,\phi}(x,y,x')\right]\!\right\}
\label{eq:10}
\end{align}
Here, $\lambda \in \{\theta,\phi\}$ and $\delta$ is the learning coefficient.

\subsection{Generator}
\label{subsec:gene}
The output $y$ of the generator is the action decided following the policy $\pi(y\!\mid\! x)$.
The reward $\hat{r}(x,y,x')$ is estimated from the output of the discriminator as follows:
\begin{equation}
\hat{r}(x,y,x') = \ln(D(x,y,x'))-\ln(1-D(x,y,x'))
\label{eq:11}
\end{equation}
The policy $\pi(y\!\mid\! x)$ of the generator is learned with any reinforcement learning method using this estimated reward.

\begin{figure}[t]
  \centering
  \includegraphics[width=\columnwidth]{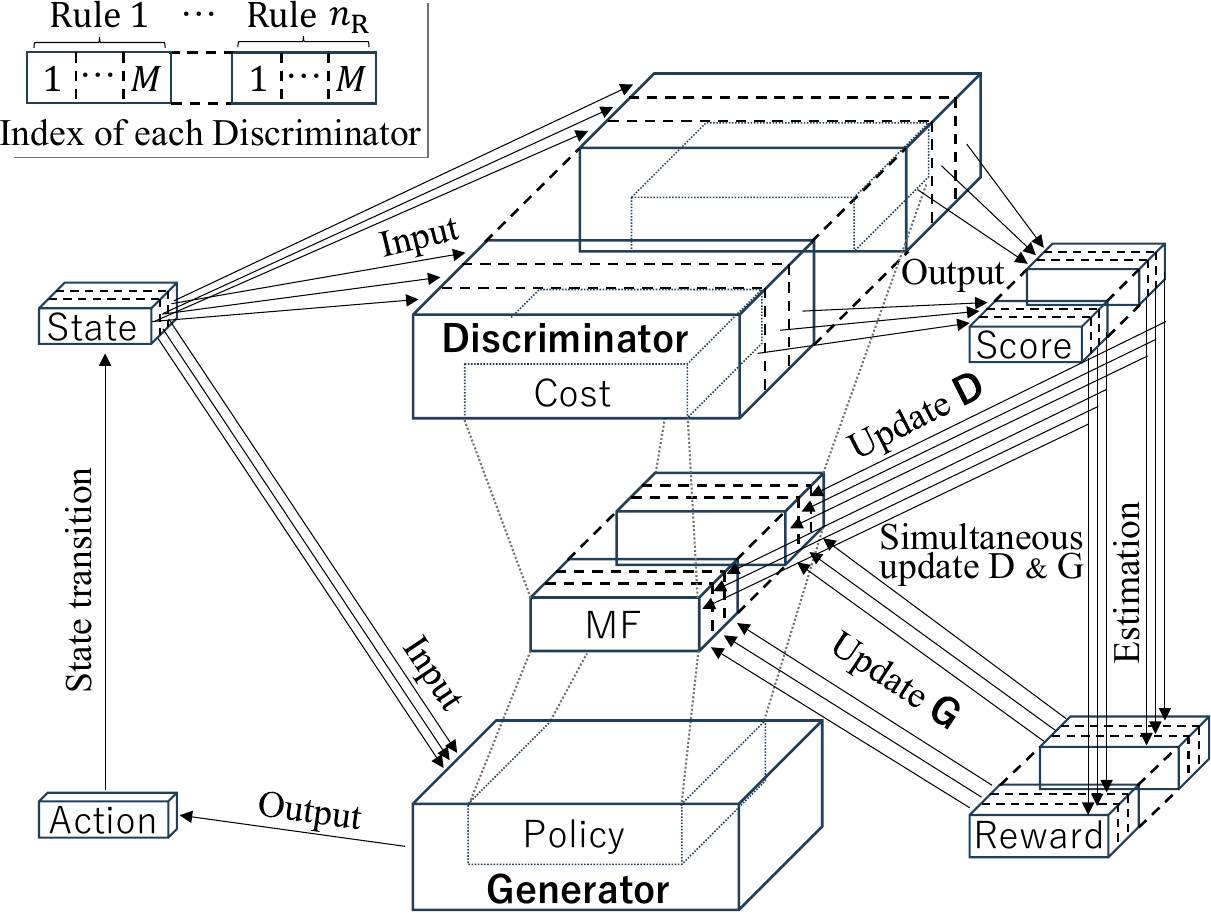}
  \caption{Configuration of the proposed method}
  \label{fig:1}
\end{figure}

\section{Proposed Method}
\label{sec:pro-meth}

The proposed method integrates the reward estimation mechanism of AIRL presented in Section~\ref{sec:AIRL} into the base model presented in Section~\ref{sec:base-mdl} by sharing the membership functions.
Fig. \ref{fig:1} shows the configuration of the proposed method.
The proposed method treats the policy of the base model as the generator and also uses the set of membership functions (MF) in the policy as the cost function of the discriminator.
Specifically, a cost function is defined using each membership function of the antecedent part, and this gives a discriminator for each fuzzy representation of each rule.
From the viewpoint of the $j$-th fuzzy representation in the antecedent part of rule $i$, the $j$-th ($=1,2,\cdots,M$) discriminator of rule $i$ thus gives a score indicating how likely a trajectory of actions and states is to be contained in the demonstration.
The outputs of the discriminators are used for two updates, one minimizing the loss function of the discriminator (Update D) and the other maximizing the expected reward of the generator (Update G), which are performed simultaneously on the parameters of the membership functions of the antecedent part shared by the discriminator and the generator.
The following subsections present the proposed cost function and discriminator, together with the learning rules for the discriminator and the generator of the proposed method.

\subsection{Cost Function and Discriminator Sharing the Membership Functions}
\label{subsec:pro-disc}

This paper proposes the following cost function defined using each membership function in the antecedent part.
\begin{align}
&f_{\alpha^i_j}(x(t),y(t),x(t+1))\nonumber\\
&\quad = \ln\Bigl\{1-\bigl|v_{\alpha^i_j}(x(t+1);x(t),y(t)) \bigr| \Bigr\}
\label{eq:12}
\end{align}
where
\begin{align}
&v_{\alpha^i_j}(x(t+1);x(t),y(t))\nonumber\\
&\quad \equiv\mathbb{E}_{\mathcal{D}_{t}}\left[ A^i_j(x'_j;\alpha^i_j)\right] -A^i_j(x_j(t+1);\alpha^i_j)
\label{eq:13}
\end{align}
Under the assumption that $0<A^i_j(x_j)<1$, it follows that $0\leq |v_{\alpha^i_j}|<1$ and $f_{\alpha^i_j}\leq 0$.
Eqs. (\ref{eq:12}) and (\ref{eq:13}) show that the value of $f_{\alpha^i_j}$ approaches zero as the matching degree $A^i_j(x_j(t+1))$ for the input approaches the expected matching degree $\mathbb{E}_{\mathcal{D}_{t}}[ A^i_j(x'_j)]$ for the demonstration.
The cost function $f_{\alpha^i_j}$ therefore evaluates the value of $(x(t),y(t))$ by the matching degree for $x_j(t+1)$.
Under the optimal policy, $\mathbb{E}_{\pi_t}[ A^i_j(x'_j)]$ equals $\mathbb{E}_{\mathcal{D}_{t}}[ A^i_j(x'_j)]$.
In this case, the difference $f_{\alpha^i_j}(x(t),y(t),x(t+1))-\mathbb{E}_{\pi_t}[f_{\alpha^i_j}(x,y,x')]$ between the value of a given $(x(t),y(t))$ and the expected value over all samples generated by $\pi$ represents the relative value of the action $y(t)$ with respect to the optimal policy $\pi$.
This difference can therefore be regarded as a form of approximation of the advantage without using a value function.
Using the cost function in Eq. (\ref{eq:12}), the output of the $j$-th discriminator of rule $i$ is defined as
\begin{align}
&D^i_j(x(t),y(t),x(t+1))\nonumber\\
& = \!\frac{\exp(f_{\!\alpha^i_j}(x(t),y(t),x(t\!+\!1)))}{\exp(f_{\!\alpha^i_j}\!(x(t),y(t),x(t\!+\!1)\!)\!)\!+\!\pi(y(t)\!\mid\! x(t)\!)}
\label{eq:14}
\end{align}

\subsection{Learning Rule for the Discriminator}
\label{subsec:LR-disc-pro}

The loss function of the discriminator in the proposed method is
\begin{align}
\mathcal{L}(\alpha^i_j) = \sum_{t=0}^{L-1}&\left\{-\mathbb{E}_{\mathcal{D}_{t}}\!\left[\ln(D^i_j(x,y,x'))\right]\right.\nonumber\\
&\ \left.-\mathbb{E}_{\pi_t}\!\left[\ln(1-D^i_j(x,y,x'))\right]\right\}
\label{eq:15}
\end{align}
In updating the discriminator, the parameters of the policy $\pi$ of the generator at time $t$ are fixed, and $\pi(y(t)\!\mid\! x(t))$ in Eq. (\ref{eq:14}) is treated as a constant.
The learning rule that minimizes this loss function is then given by
\begin{align}
\Delta\alpha^i_j\! = \delta\!\sum_{t=0}^{L-1}&\!\!\left\{\!\mathbb{E}_{\mathcal{D}_{t}}\!\!\left[\!(1\!-\!D^i_j(x,y,x')\!)\frac{\partial}{\partial\alpha^i_j}f_{\!\alpha^i_j}\!(x,y,x')\!\right]\right.\nonumber\\
&\ \left.-\mathbb{E}_{\pi_t}\!\!\left[\!D^i_j(x,y,x')\frac{\partial}{\partial\alpha^i_j}f_{\alpha^i_j}(x,y,x')\!\right]\!\right\}
\label{eq:16}
\end{align}
The gradient used for this minimization is thus only that of the cost function $f_{\alpha^i_j}$.
Even though the policy $\pi$ of the generator shares the parameters with the discriminator, it is treated as a constant in the learning rule for the discriminator.
The derivation of Eq. (\ref{eq:16}) is given in Appendix~\ref{app:dev16}.

\subsection{Learning Rule for the Generator}
\label{subsec:LR-gene-pro}

The evaluation function of the generator in the proposed method, which the policy gradient method maximizes, is defined as
\begin{equation}
J(\mu) = \mathbb{E}_{\tau\sim\pi}\!\left[\sum_{t=0}^{L-1}\gamma^t\hat{r}(x(t),y(t),x(t+1))\right]
\label{eq:17}
\end{equation}
where
\begin{align}
&\hat{r}(x,y,x') = \left\{
\begin{array}{ll}
\! \hat{r}^i_j(x,y,x') & \textrm{if}\ \ \mu=\alpha^i_j\\[2mm]
\! \sum_{j=1}^{M}\hat{r}^i_j(x,y,x') & \textrm{if}\ \ \mu\in\{\beta^i_k, \eta_i\}
\end{array}
\right.
\label{eq:18}\\
&\hat{r}^i_j(x,y,x') = \ln(D^i_j(x,y,x'))-\ln(1\!-\!D^i_j(x,y,x'))
\label{eq:19}
\end{align}
Here, $\gamma$ is the discount rate.
In updating the generator, the parameters of the discriminators $D^i_j$ at time $t$ are fixed, and $\hat{r}(x(t),y(t),x(t+1))$ in Eq. (\ref{eq:17}) is treated as a constant.
The learning rule that maximizes this evaluation function is then given by
\begin{align}
&\Delta\mu\nonumber\\
&=\! \varepsilon \mathbb{E}_{\tau\sim\pi}\!\!\left[\sum_{t=0}^{L-1}\!\!\left\{\!\sum_{u=q}^{L-1}\!\gamma^{u-q}\hat{r}(x(\!u\!),y(\!u\!),x(\!u+1\!)\!)\!\!\right\}\!e_{\mu}(t)\!\right]
\label{eq:20}
\end{align}
Here, $q=0$ corresponds to Eq. (\ref{eq:4}) and $q=t$ corresponds to the REINFORCE learning rule.
The gradient used for this maximization is thus only the characteristic eligibility $e_{\mu}$.
Even though the reward $\hat{r}$ is given by the outputs of the discriminators $D^i_j$, which share the parameters with the generator, it is treated as a constant in the learning rule for the generator.

With the learning rule in Eq. (\ref{eq:20}), the parameters of the membership functions in the consequent part and of the rule weights can also be learned, in addition to those in the antecedent part.
In the following discussion, however, only the parameters $\alpha^i_j$ of the membership functions in the antecedent part shared with the discriminator are considered as the learning target in the generator.
The characteristic eligibility $e_{\alpha^i_j}(t)$ for the update $\Delta\alpha^i_j$ is given by
\begin{align}
&e_{\alpha^i_j}(t)\nonumber\\
&=\! -\frac{1}{T}\eta_i \!\left\{\mathbb{E}_{\pi(y\mid x(t)\!)}\!\!\left[B^i(y)\right]\!-\!B^i(y(t))\!\right\}\!\frac{\partial}{\partial \alpha^i_j}A^i(x(t))
\label{eq:21}
\end{align}
The derivation of Eq. (\ref{eq:21}) is given in Appendix~\ref{app:dev21}.

\subsection{Simultaneous Learning}
\label{subsec:simul}

In the proposed method, the discriminator and the generator share the parameters, and $\alpha^i_j$ is updated by Eqs. (\ref{eq:16}) and (\ref{eq:20}) simultaneously.
A single update with one set of samples therefore covers both the discriminator and the reward estimation for the generator, just as one alternating cycle does in AIRL.
In this case, the amount of a single update of $\alpha^i_j$ is given by
\begin{align}
&\Delta\alpha^i_j\! =\! \delta\!\sum_{t=0}^{L-1}\!\!\left\{\!\mathbb{E}_{\mathcal{D}_{t}}\!\!\left[\!(1\!-\!D^i_j(x,y,x')\!)\frac{\partial}{\partial\alpha^i_j}f_{\!\alpha^i_j}(x,y,x')\!\right]\right.\nonumber\\
&\quad\quad\quad\quad\quad\left.-\mathbb{E}_{\pi_t}\!\!\left[\!D^i_j(x,y,x')\frac{\partial}{\partial\alpha^i_j}f_{\alpha^i_j}(x,y,x')\!\right]\!\right\}\nonumber\\
&+\!\varepsilon \mathbb{E}_{\tau\sim\pi}\!\!\left[\sum_{t=0}^{L-1}\!\!\left\{\!\sum_{u=q}^{L-1}\!\gamma^{u-q}\hat{r}(x(\!u\!),y(\!u\!),x(\!u+1\!)\!)\!\!\right\}\!e_{\alpha^i_j}\!(\hspace{-0.05em}t\hspace{-0.05em})\!\right]
\label{eq:22}
\end{align}

In Eq. (\ref{eq:22}), the first term on the right-hand side corresponds to an update that increases the difference in the output of the membership function $A^i_j$ between the demonstration and the sample, while the second term corresponds to an update that decreases it.
If the update of the generator can bring the shapes of the membership functions close to a constant function, the output of the cost function easily becomes large regardless of the state of the sample, and a policy that generates trajectories similar to the demonstration does not necessarily have to be learned in the generator.
Because the first term keeps the membership function away from such a constant-like shape, the second term can update the membership function so that the policy generates samples close to the states of the demonstration.

\section{Control Task and Its Application}
\label{sec:task}

This section describes a simulation problem \cite{IshiharaMF2024} that assumes an illustrative vehicle speed control task, and the application of the proposed method to this problem.

\subsection{Vehicle Speed Control Problem}
\label{subsec:VSC}

Two vehicles are assumed to travel in the same direction along a one-dimensional straight line representing a road, with the speed of the leading vehicle $v_{\textrm{front}}$ (km/h) held constant.
The task of this problem is to control the speed of the following vehicle $x_{\textrm{v}}$ (km/h) by accelerating and braking, so that the inter-vehicle distance $x_{\textrm{d}}$ (m) enters the target range $[l_{\textrm{near}},l_{\textrm{far}}]$ (m) within 80 seconds from the initial state at $t=0$ (sec) and then remains within that range until $t=110$ (sec).

With the time discretized at $\Delta t=1$ (sec), the speed control of the following vehicle starts at $t=0$ and ends when $t=110$ is reached.
The length of one episode is therefore $L=110$.
However, if the inter-vehicle distance becomes less than 0 m, which means a rear-end collision, or reaches 400 m or more, the control is terminated at that point and the episode is regarded as ended.

\subsection{Fuzzy Inference Rules}
\label{subsec:VSC-rule}

The fuzzy inference rules for controlling the speed of the following vehicle are given by Eq. (\ref{eq:1}) with $M=2$ and $N=1$.
The input variables in the antecedent part are $\boldsymbol{x}_1=(x_{\textrm{d}},l_{\textrm{near}},l_{\textrm{far}})$ and $\boldsymbol{x}_2=(x_{\textrm{v}},v_{\textrm{front}})$.
The output variable $y_1$ in the consequent part corresponds to the accelerator input ($0\le y_1 \le 5$) and the brake input ($-5\le y_1 \le 0$), and the resulting acceleration of the following vehicle is $2y_1$ (km/h/sec).

The fuzzy sets $A_1^{\textrm{long}}$ and $A_1^{\textrm{short}}$ correspond to fuzzy representations of the inter-vehicle distance, long and short, respectively.
The fuzzy sets $A_2^{\textrm{fast}}$ and $A_2^{\textrm{slow}}$ correspond to fuzzy representations of the speed of the following vehicle, fast and slow, respectively.
The fuzzy sets $B_1^{\textrm{high-ac.}}$, $B_1^{\textrm{low-ac.}}$, $B_1^{\textrm{high-br.}}$, $B_1^{\textrm{low-br.}}$, and $B_1^{\textrm{no-op.}}$ correspond to fuzzy representations of high accelerator input (ac.), low accelerator input, high brake input (br.), low brake input, and no operation (no-op.), respectively.
With these combinations ($2\times 2\times 5=20$), there are 20 fuzzy inference rules and rule weights $\eta_i$ for $i=1$ to $i=20$.

\subsection{Membership Functions and Parameters}
\label{subsec:MF}

The membership functions for the nine fuzzy sets are defined as
\begin{align}
A^i_1(\boldsymbol{x}_1;\alpha^i_1)\!\equiv\! \left\{
\begin{array}{ll}
\!\!A^{\textrm{long}}_1(\boldsymbol{x}_1) & \textrm{if}\ \ i=1,...,10\\[2mm]
\!\!A^{\textrm{short}}_1(\boldsymbol{x}_1) & \textrm{if}\ \ i=11,...,20
\end{array}
\right.
\label{eq:23}
\end{align}
\begin{align}
A^i_2(\boldsymbol{x}_2;\alpha^i_2)\!\equiv\! \left\{
\begin{array}{ll}
\!\!A^{\textrm{fast}}_2(\boldsymbol{x}_2) & \textrm{if}\ \ i=1,...,5,11,...,15\\[2mm]
\!\!A^{\textrm{slow}}_2(\boldsymbol{x}_2) & \textrm{if}\ \ i=6,...,10,16,...,20\\
\end{array}
\right.
\label{eq:24}
\end{align}
\begin{align}
B_1^i(y_1;\beta^i_1)\! \equiv\! \left\{
\begin{array}{ll}
\!\!B_1^{\textrm{high-ac.}}(y_1) & \textrm{if}\ \ i=1,6,11,16\\[2mm]
\!\!B_1^{\textrm{low-ac.}}(y_1) & \textrm{if}\ \ i=2,7,12,17\\[2mm]
\!\!B_1^{\textrm{high-br.}}(y_1) & \textrm{if}\ \ i=3,8,13,18\\[2mm]
\!\!B_1^{\textrm{low-br.}}(y_1) & \textrm{if}\ \ i=4,9,14,19\\[2mm]
\!\!B_1^{\textrm{no-op.}}(y_1) & \textrm{if}\ \ i=5,10,15,20\\
\end{array}
\right.
\label{eq:25}
\end{align}
A common membership function is thus used for the same fuzzy set even in different rules.
For example, the membership functions for the fuzzy sets $A_1^1$ through $A_1^{10}$ are all $A^{\textrm{long}}_1(\boldsymbol{x}_1)$, which corresponds to the fuzzy representation ``long.''

\subsection{Deciding the Action}
\label{subsec:dec-a}

In the training phase, the output value for the action at time $t$ is selected probabilistically from the discrete values of $y_1$ according to the policy $\pi(y_1\!\mid\! \boldsymbol{x}(t))$.
In the evaluation phase, the output value is decided deterministically as the expected value $\sum_{y_1}y_1\pi (y_1\!\mid\! \boldsymbol{x}(t))$ over the discrete values of $y_1$.

\begin{figure}[t]
  \centering
  \includegraphics[width=\columnwidth]{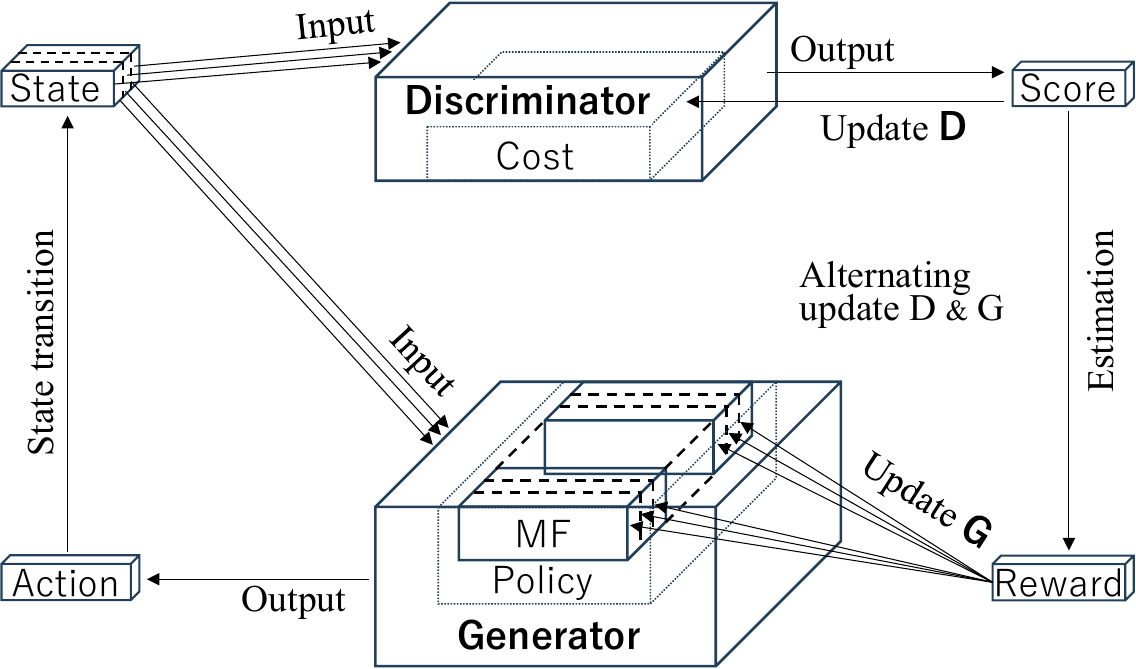}
  \caption{Configuration of AIRL-BMP}
  \label{fig:2}
\end{figure}

\section{Computer Experiments}
\label{sec:exp}

In the computer experiments, the parameters were learned by applying the proposed method to the control task described in Section~\ref{sec:task}, and the results were evaluated.
For comparison, experiments were also conducted with AIRL-BMP (AIRL with Base-Model Policy), an IRL method that uses the discriminator of AIRL described in Section~\ref{sec:AIRL} as it stands and takes the policy of the base model as the generator.
Fig. \ref{fig:2} shows the configuration of AIRL-BMP.
In AIRL-BMP, the discriminator and the generator share no parameters, and they are updated alternately rather than simultaneously.
The learning rule for the discriminator in AIRL-BMP is Eq. (\ref{eq:10}), and that for the generator is given by substituting Eq. (\ref{eq:11}) into $\hat{r}$ in Eq. (\ref{eq:20}), the learning rule of the proposed method.

In these experiments, each membership function in the antecedent part, namely $A^{\textrm{long}}_1(\boldsymbol{x}_1)$, $A^{\textrm{short}}_1(\boldsymbol{x}_1)$, $A^{\textrm{fast}}_2(\boldsymbol{x}_2)$, and $A^{\textrm{slow}}_2(\boldsymbol{x}_2)$, was represented by a neural network.
Among the parameters $\{\alpha,\beta,\eta\}$ of the policy, the weight parameters $\alpha^i_j\in\{\alpha^{\textrm{long}}_1, \alpha^{\textrm{short}}_1, \alpha^{\textrm{fast}}_2, \alpha^{\textrm{slow}}_2\}$ of these neural networks were the learning target.
The membership functions $B_1^{i}(y_1)$ in the consequent part were fixed to those adopted in the previous experiments on the base model \cite{IshiharaMF2024}.
The definitions of $B_1^{i}(y_1)$ and the discretization of $y_1$ are given in Appendix~\ref{app:mfc}.
The rule weights $\eta_i$ were fixed as
\begin{align}
\eta_i = \left\{
\begin{array}{ll}
\!\! 1 & \textrm{if}\ \ i=5,6,13,20\\
\!\! 0 & \textrm{otherwise}\ \ 
\end{array}
\right.
\label{eq:26}
\end{align}
In Eq. (\ref{eq:26}), for each of the four combinations of antecedent fuzzy representations, a weight of 1 was assigned to the fuzzy inference rule whose consequent fuzzy representation was considered the most appropriate, and 0 to the others.

For both methods, the temperature $T$ in Eq. (\ref{eq:3}) for the policy was set to $0.02$.
In Eq. (\ref{eq:20}) for the learning rule of the generator, $q$ and $\gamma$ were set to $0$ and $1$, respectively.
For the proposed method, the learning coefficients $\delta$ in Eq. (\ref{eq:16}) and $\varepsilon$ in Eq. (\ref{eq:20}) were set to $0.00002$ and $0.000001$, respectively.
For AIRL-BMP, the learning coefficients $\delta$ in Eq. (\ref{eq:10}) and $\varepsilon$ in Eq. (\ref{eq:20}) were set to $0.0002$ and $0.00003$, respectively.
The discount rate $\kappa$ in Eq. (\ref{eq:8}) for the cost function of AIRL-BMP was set to $0.9$.
These values were chosen empirically based on the results of preliminary experiments.

\subsection{Configuration of the Neural Networks}
\label{subsec:NN}

\subsubsection{Membership Functions in the Antecedent Part}
\label{subsubsec:MF-AP}

Each of the neural networks representing the four membership functions in the antecedent part consisted of an input layer with three or two units, a hidden layer with 120 units, and an output layer with one unit that outputs the matching degree.
The activation functions were PReLU in the hidden layer and the sigmoid function in the output layer.
The initial negative slope of PReLU was $0.25$.
The weight parameters corresponding to the biases were initialized with a uniform distribution over $[0,0.01]$, and the other weight parameters were initialized with a normal distribution of standard deviation $1/\sqrt{n}$, where $n$ is the number of input units of each layer.
The input values to the neural networks were scaled to one hundredth of the original values, namely $(x_{\textrm{v}},v_{\textrm{front}})/100$ or $(x_{\textrm{d}},l_{\textrm{near}},l_{\textrm{far}})/100$.

\subsubsection{Discriminator of AIRL-BMP}
\label{subsubsec:disc-AIRL}

In the discriminator of AIRL-BMP, the input of the function $g(\boldsymbol{x}_1,\boldsymbol{x}_2,y_1;\theta)$ was six-dimensional and that of the function $h(\boldsymbol{x}_1,\boldsymbol{x}_2;\phi)$ was five-dimensional.
Each of these functions was represented by a neural network consisting of an input layer with six or five units, a hidden layer with 120 units, and an output layer with one unit.
The activation functions were PReLU in the hidden layer and the identity function in the output layer.
The initial negative slope of PReLU, the initial values of the weight parameters, and the scaling of $\boldsymbol{x}_1$ and $\boldsymbol{x}_2$ were the same as in the neural networks representing the membership functions in the antecedent part.
In addition, $y_1$ was scaled to one fifth of the original value when it was input to the network.

\begin{table*}[t]
\centering
\begin{threeparttable}
\caption{Control results for the 580 training problems (task success rate, number of successful tasks, and breakdown of failed tasks) over the 20 sets of learned parameters}\label{tab:train}
\tablebodyfont
\setlength{\tabcolsep}{4pt}
\begin{tabular}{@{}lcccccccc@{}}
\toprule
 & Avg. task
 & \multicolumn{4}{c}{Successful tasks}
 & \multicolumn{3}{c}{Avg. no. of failed tasks} \\
\cmidrule(lr){3-6}\cmidrule(l){7-9}
 & success rate & min. & max. & avg. & SD
 & collision & out of range & delay \\
\midrule
Proposed   & 0.981 & 565 & 572 & 569.25 & \,\ 2.34 & 10.50 & \,\ 0.00 & \,\ 0.25 \\
AIRL-BMP   & 0.721 & 285 & 539 & 418.45 & 74.59 & 19.45 & 44.15 & 97.95 \\
\midrule
Base model\tnote{a} & 0.983 & 519 & 580 & 570.05 & 12.16 & \,\ 6.90
           & \multicolumn{2}{c}{3.05\tnote{b}} \\
\botrule
\end{tabular}
\begin{tablenotes}
\item[a] Values quoted from \cite{IshiharaMF2024}
\item[b] Combined count for ``out of range'' and ``delay'' failures, which are not distinguished in the base model
\end{tablenotes}
\end{threeparttable}
\end{table*}
\begin{table}[t]
\caption{Number of training epochs over the 20 training runs}\label{tab:epochs}
\begin{tabular}{@{}lcccc@{}}
\toprule
 & \multicolumn{4}{c}{Training epochs} \\
\cmidrule(l){2-5}
 & min. & max. & avg. & SD \\
\midrule
Proposed & \,\ 35 & \,\ \,\ 78 & \,\ 48.80 & \,\ 10.51 \\
AIRL-BMP & 327 & 1350 & 746.65 & 257.83 \\
\botrule
\end{tabular}
\end{table}

\subsection{Experimental Procedure in the Training Phase}
\label{subsec:exp-proc}

\subsubsection{Training Problems}
\label{subsubsec:train-prob}

A total of $5\times 5\times 5\times 5=625$ speed control problems were prepared by exhaustively combining the speed of the leading vehicle (km/h) from $\{10,30,50,70,90\}$, the initial speed of the following vehicle (km/h) from $\{0,20,40,60,80\}$, the initial inter-vehicle distance (m) from $\{10,30,50,70,90\}$, and the target range of the inter-vehicle distance (m) from $\{ [10,15],[30,35],\allowbreak [50,55],[70,75],[90,95]\}$.
Problems in which a rear-end collision was theoretically unavoidable or hard to prevent were then excluded: those in which the initial speed of the following vehicle exceeded the speed of the leading vehicle by 50 km/h or more and the initial inter-vehicle distance was 30 m or less, and those in which the initial speed of the following vehicle exceeded the speed of the leading vehicle by 70 km/h or more.
The remaining 580 problems were used for training.
For each of these 580 problems, one successful episode of the demonstration $\tau_{\textrm{D}}$ was created by manual control and used as the expert trajectory.

\subsubsection{Training Procedure}
\label{subsubsec:train-proc}

After the parameters to be learned were initialized, the following procedure was used.
\begin{description}
  \item[Step 1.]Generate a trajectory sample $\tau_{\textrm{S}}$ for a training problem from the output of the generator, using the action decision in the training phase.\vspace{1mm}
  \item[Step 2.]For each training problem, repeat the sample generation of Step 1 and the update below in sequence. Proceed to Step 3 when the parameters have been updated once for each training problem.
    \begin{list}{}{
    \setlength{\leftmargin}{1em}   
    \setlength{\labelwidth}{0pt}   
    \setlength{\itemindent}{0.5em}  
    \let\makelabel\textit          
  }
    \item[Proposed method:]Update the parameters $\alpha^i_j$ shared by the discriminator and the generator using the learning rule in Eq. (\ref{eq:22}).\vspace{0.5mm}
    \item[AIRL-BMP: ]Update the parameters $\{\theta,\phi\}$ of the discriminator using the learning rule in Eq. (\ref{eq:10}). Then update the parameters $\alpha^i_j$ of the generator using the learning rule in Eq. (\ref{eq:20}), into which the reward in Eq. (\ref{eq:11}) from the updated discriminator is substituted.
  \end{list}\vspace{1mm}
  \item[Step 3.]Generate a trajectory sample $\tau_{\textrm{S}}$ for a training problem from the output of the generator, using the action decision in the evaluation phase, then check and record whether the task succeeds. Proceed to Step 4 when this has been done for all training problems.\vspace{1mm}
  \item[Step 4.]If the task succeeds for all training problems, adopt the current parameters as the result and end the training. If any problem fails, return to Step 1. However, the number of epochs, each consisting of Steps 1 to 4, is limited to 500 for the proposed method and 2500 for AIRL-BMP. When this maximum is reached, adopt as the result the parameters that gave the largest number of successful problems, and end the training. If more than one set of parameters qualifies, adopt the one obtained at the smallest number of epochs.
\end{description}
For both methods, stochastic gradient descent (SGD) was used as the optimizer. The update equations derived from each learning rule are given in Appendix~\ref{app:upd-eqs}.

The procedure described above, from the initialization of the parameters to the end of the training, is counted as one training run.
Each training run therefore gives one set of learned parameters.
In this study, 20 training runs were conducted, giving 20 sets of learned parameters.

\subsection{Experimental Results and Discussion in the Training Phase}
\label{subsec:result-train}

The control results for the training problems with the weight parameters from each training run are shown in Table~\ref{tab:train}, and the number of epochs for each run is shown in Table~\ref{tab:epochs}.
For comparison, Table~\ref{tab:train} also includes the control results reported in \cite{IshiharaMF2024} for the same training problems with the base model, which uses a reward function set manually in advance.
Table~\ref{tab:train} shows that the proposed method achieved an average task success rate of $98.1\%$, close to the $98.3\%$ of the base model, with small variation across the 20 sets of results.
The average task success rate of AIRL-BMP, in contrast, was $72.1\%$, about $26$ points lower than that of the proposed method.
Moreover, AIRL-BMP gave a smaller minimum and a smaller maximum number of successfully controlled problems than the proposed method, and its variation across the sets was also greater, with a standard deviation about 32 times as large.
Table~\ref{tab:epochs} shows that the number of epochs required for training by the proposed method was about one fifteenth of that for AIRL-BMP, and that its variation across training runs also tended to be small, with a standard deviation about one twenty-fifth as large.
A Mann-Whitney U test between the proposed method and AIRL-BMP showed that the proposed method achieved a significantly larger number of successful tasks (median $570.00$, IQR $[567.75, 571.25]$ versus median $407.50$, IQR $[370.50, 478.75]$; $U=400$, $p<0.001$, $r=0.855$). It also required significantly fewer epochs (median $47.00$, IQR $[40.75, 55.00]$ versus median $662.50$, IQR $[584.25, 887.25]$; $U=0$, $p<0.001$, $r=0.855$), with $n=20$ for each method\footnote{The statistical test is applied only to the results obtained in the experiments of this paper.}.

Task failures can be classified into three types: collision, being out of the target range at $t=110$ (out of range), and remaining within the target range until $t=110$ with the start of that period later than $t=80$ (delay).
The breakdown of the average number of failed tasks in Table~\ref{tab:train} shows that the proposed method had no cases of ``out of range'' and only 0.25 cases of ``delay,'' and the base model also had only 3.05 cases of ``out of range'' and ``delay'' combined.
AIRL-BMP, in contrast, had a large total average number of failed tasks, with ``delay'' the most frequent at 97.95 cases, followed by ``out of range'' at 44.15 cases.
These results suggest that during training AIRL-BMP could not improve many trajectories of ``delay'' and ``out of range,'' which are relatively close to those of successful tasks, whereas the proposed method was able to learn to succeed in these tasks.
Most of the 10.50 cases of ``collision'' in the proposed method occurred in relatively demanding problem settings in which the initial speed of the following vehicle was 50 km/h higher than the speed of the leading vehicle, as was also the case for ``collision'' in the base model.

\begin{figure}[tb]
  \centering
  \includegraphics[width=\columnwidth]{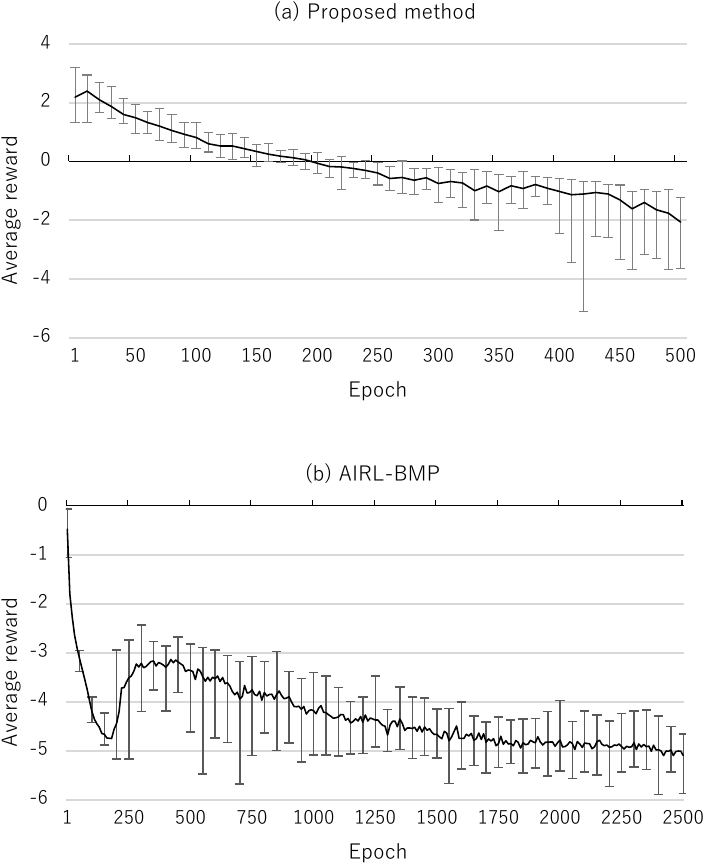}
  \caption{Average reward per time step for all fuzzy representations at each training epoch. \textbf{a} Proposed method, \textbf{b} AIRL-BMP}
  \label{fig:3}
\end{figure}
\begin{table*}[t]
\centering
\begin{threeparttable}
\caption{Trajectory characteristics for the successful training problems}\label{tab:chara}
\tablebodyfont
\setlength{\tabcolsep}{4pt}
\begin{tabular}{@{}lccccccc@{}}
\toprule
 & \multicolumn{2}{c}{Time to settle}
 & \multicolumn{2}{c}{Settling delay}
 & \multicolumn{2}{c}{Exits from range}
 & SD after settling \\
\cmidrule(lr){2-3}\cmidrule(lr){4-5}\cmidrule(lr){6-7}\cmidrule(l){8-8}
 & avg. & SD & avg. & SD & avg. & SD &pooled value\\
\midrule
Proposed & 22.83 & 1.84 & \,\ 5.71 & \,\ 1.28 & 0.48 & 0.13 & 0.28\\
AIRL-BMP & 47.91 & 6.65 & 29.51 & 10.22 & 1.35 & 0.41 & 0.96\\
\midrule
Demonstration & 14.60 & --- & \,\ 2.68 & --- & 0.21 & --- & 0.35\\
\botrule
\end{tabular}
\end{threeparttable}
\end{table*}
\begin{figure}[t]
  \centering
  \includegraphics[width=\columnwidth]{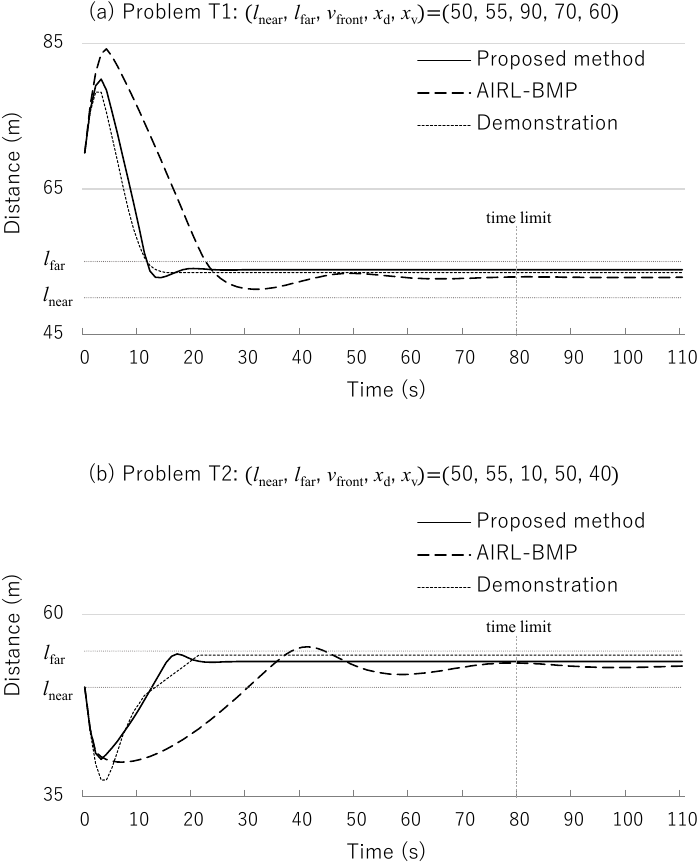}
  \caption{Trajectories of the inter-vehicle distance with the proposed method, AIRL-BMP, and the demonstration. \textbf{a} Problem T1, \textbf{b} Problem T2}
  \label{fig:4}
\end{figure}

The average reward per time step for all fuzzy representations at each training epoch is shown in Fig. \ref{fig:3}.
The value at each epoch is averaged over all the training problems in all the training runs, and the error bars show the minimum and maximum of the averages over the 580 problems in each training run.
Fig. \ref{fig:3} shows that the average reward for the proposed method decreased almost monotonically toward zero, the theoretical value of the reward under the optimal policy, within a small number of epochs.
This suggests that the proposed method efficiently learned parameters achieving a high average task success rate.
The average reward for AIRL-BMP, in contrast, started from a negative value, dropped sharply, and then rose sharply once. 
It never reached zero in any of the training runs, however, and thereafter decreased almost monotonically.
When the trajectories of the samples from the generator differ from those of the demonstration, with large deviations in the inter-vehicle distance or the speed, the output of the discriminator generally becomes small and the reward tends to take a large negative value.
Since the average reward for AIRL-BMP never became zero or higher throughout training with five times as many epochs as for the proposed method, AIRL-BMP was less effective than the proposed method in learning to bring the trajectories of the samples for each problem closer to those of the demonstration on average.

As successful examples among the control results for the training problems, the trajectories of the inter-vehicle distance after training with the proposed method and with AIRL-BMP are shown in Fig. \ref{fig:4}, together with the trajectory of the demonstration, for problem T1, which required acceleration from the initial state, and problem T2, which required deceleration.
In problems T1 and T2, $(l_{\textrm{near}},l_{\textrm{far}},v_{\textrm{front}})$ were $(50,55,90)$ and $(50,55,10)$, and the initial values of $(x_{\textrm{d}},x_{\textrm{v}})$ were $(70,60)$ and $(50,40)$, respectively.
For both methods, the trajectories in Fig. \ref{fig:4} were generated by the policies using the parameters from the training run that gave the largest number of successful problems.
Fig. \ref{fig:4} shows that the proposed method, as in the demonstration, tended to settle the inter-vehicle distance within the target range earlier than AIRL-BMP and to give trajectories with smaller oscillations.
The trajectories from AIRL-BMP, in contrast, tended to settle within the target range later and to require more time for the oscillations to converge.
Table~\ref{tab:chara} quantifies these characteristics for the successful training problems.
The time to settle is the start of the period in which the inter-vehicle distance remained within the target range until $t=110$, the settling delay is the interval between the first time the distance was within the range and the time to settle, the exits from range is the number of times the distance moved from inside to outside the target range, and the SD after settling is the standard deviation of the distance from the time to settle to $t=110$.
For the first three measures, the values were averaged over the successful problems for each set of learned parameters, and Table~\ref{tab:chara} shows the mean and the standard deviation of these averages over the 20 sets.
The last measure is pooled over all the successful problems in all the sets.
Table~\ref{tab:chara} also includes the values for the demonstration for reference. 
Since only one demonstration was created for each problem, the values were averaged over the 580 problems and no standard deviation over sets is available.
A Mann-Whitney U test between the proposed method and AIRL-BMP showed that the differences in all four measures were significant ($p<0.001$, $n=20$ for each method), where the pooled standard deviation for each set was used for the last measure.
Continuous oscillations of the trajectory do not occur in the demonstration and represent unnatural control.
They not only increase the possibility of task failure but can also reduce usefulness even when the task succeeds, and they are generally undesirable.
For the proposed method, in contrast, such oscillations requiring time to converge hardly occurred.
These results indicate that the proposed method was less prone to continuous oscillations of the trajectory than AIRL-BMP.

\begin{table*}[tb]
\centering
\begin{threeparttable}
\caption{Control results for the 722 evaluation problems (task success rate, number of successful tasks, and breakdown of failed tasks) over the 20 sets of learned parameters}\label{tab:eval}
\tablebodyfont
\setlength{\tabcolsep}{4pt}
\begin{tabular}{@{}lcccccccc@{}}
\toprule
 & Avg. task
 & \multicolumn{4}{c}{Successful tasks}
 & \multicolumn{3}{c}{Avg. no. of failed tasks} \\
\cmidrule(lr){3-6}\cmidrule(l){7-9}
 & success rate & min. & max. & avg. & SD
 & collision & out of range & delay \\
\midrule
Proposed   & 1.000 & 722 & 722 & 722.00 & \,\ 0.00 & 0.00 & \,\ 0.00 & \,\ 0.00 \\
AIRL-BMP   & 0.924 & 586 & 722 & 667.15 & 47.05 & 0.00 & 12.65 & 42.20 \\
\midrule
Base model\tnote{a} & 0.994 & 637 & 722 & 717.65 & 18.99 & 0.00 & \multicolumn{2}{c}{4.35\tnote{b}} \\
\botrule
\end{tabular}
\begin{tablenotes}
\item[a] Values quoted from \cite{IshiharaMF2024}
\item[b] Combined count for ``out of range'' and ``delay'' failures, which are not distinguished in the base model
\end{tablenotes}
\end{threeparttable}
\end{table*}

\subsection{Experimental Procedure in the Evaluation Phase}
\label{subsec:eval-proc}

As speed control problems different from those used for training, $5\times 5\times 5\times 5=625$ problems were prepared by exhaustively combining the speed of the leading vehicle (km/h) from $\{45,55,65,75,85\}$, the initial speed of the following vehicle (km/h) from $\{0,10,30,50,70\}$, the initial inter-vehicle distance (m) from $\{30,45,65,70,80\}$, and the target range of the inter-vehicle distance (m) from $\{[10,30],[20,40],[40,60],[50,60],[60,70]\}$.
In addition, $3\times 3\times 3\times 3=81$ problems were prepared by exhaustively combining the speed of the leading vehicle (km/h) from $\{40,50,60\}$, the initial speed of the following vehicle (km/h) from $\{20,40,60\}$, the initial inter-vehicle distance (m) from $\{20,40,60\}$, and the target range of the inter-vehicle distance (m) from $\{[10,20],[40,50],[45,60]\}$.
A further $4\times 1\times 2\times 2=16$ problems were prepared by exhaustively combining the speed of the leading vehicle (km/h) from $\{20,30,50,60\}$, the initial speed of the following vehicle (km/h) from $\{30\}$, the initial inter-vehicle distance (m) from $\{10,50\}$, and the target range of the inter-vehicle distance (m) from $\{[10,15],[30,45]\}$.
For these 722 evaluation problems, one set of control results was obtained with each of the 20 sets of learned weight parameters and under the action decision in the evaluation phase.

\begin{table*}[t]
\centering
\begin{threeparttable}
\caption{Trajectory characteristics for the successful evaluation problems}\label{tab:chara-eva}
\tablebodyfont
\setlength{\tabcolsep}{4pt}
\begin{tabular}{@{}lccccccc@{}}
\toprule
 & \multicolumn{2}{c}{Time to settle}
 & \multicolumn{2}{c}{Settling delay}
 & \multicolumn{2}{c}{Exits from range}
 & SD after settling \\
\cmidrule(lr){2-3}\cmidrule(lr){4-5}\cmidrule(lr){6-7}\cmidrule(l){8-8}
 & avg. & SD & avg. & SD & avg. & SD &pooled value\\
\midrule
Proposed & 17.51 & 2.06 & \,\ 4.56 & 0.47 & 0.28 & 0.02 & 0.81\\
AIRL-BMP & 31.26 & 3.22 & 11.61 & 3.66 & 0.45 & 0.19 & 2.20\\
\botrule
\end{tabular}
\end{threeparttable}
\end{table*}

\subsection{Experimental Results and Discussion in the Evaluation Phase}
\label{subsec:result-eval}

The control results for the evaluation problems with the weight parameters from each training run are shown in Table~\ref{tab:eval}.
For comparison, as in Table~\ref{tab:train}, Table~\ref{tab:eval} also includes the results reported in \cite{IshiharaMF2024} for the same evaluation problems with the base model.
Table~\ref{tab:eval} shows that the proposed method achieved an average task success rate of $100\%$, higher than the $99.4\%$ of the base model, with no variation across the sets.
AIRL-BMP, in contrast, had an average task success rate of $92.4\%$, lower than that of the base model, and its variation across the sets was relatively large, as was also the case for the training problems.
A Mann-Whitney U test between the proposed method and AIRL-BMP showed that the former achieved a significantly larger number of successful tasks (median $722.00$, IQR $[722.00, 722.00]$ versus median $671.50$, IQR $[638.50, 715.25]$; $U=360$, $p<0.001$, $r=0.684$, $n=20$ for each method).

In the breakdown of the average number of failed tasks for AIRL-BMP, ``delay'' was the most frequent at 42.20 cases, followed by ``out of range'' at 12.65 cases.
As was also the case for the training problems, a tendency toward insufficient learning performance was suggested.
For all three methods, no case of ``collision'' occurred in any of the results.
This is thought to be partly because there was no evaluation problem in which the initial speed of the following vehicle exceeded the speed of the leading vehicle by 50 km/h or more.

\begin{figure}[t]
  \centering
  \includegraphics[width=\columnwidth]{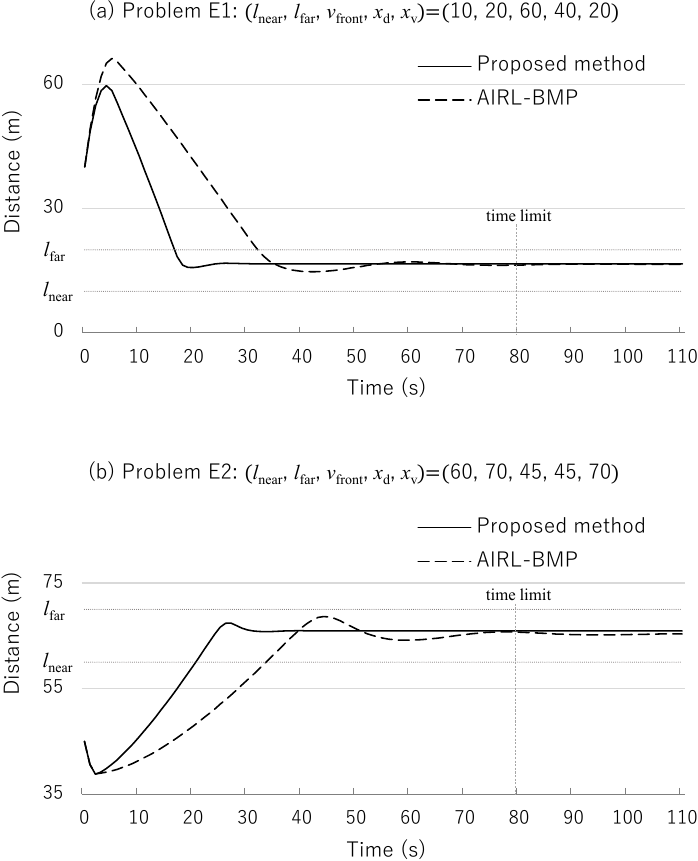}
  \caption{Trajectories of the inter-vehicle distance with the proposed method and AIRL-BMP. \textbf{a} Problem E1, \textbf{b} Problem E2}
  \label{fig:5}
\end{figure}

As successful examples among the control results for the evaluation problems, the trajectories of the inter-vehicle distance after training with the proposed method and with AIRL-BMP are shown in Fig. \ref{fig:5}, for problem E1, which required acceleration from the initial state, and problem E2, which required deceleration.
In problems E1 and E2, $(l_{\textrm{near}},l_{\textrm{far}},v_{\textrm{front}})$ were $(10,20,60)$ and $(60,70,45)$, and the initial values of $(x_{\textrm{d}},x_{\textrm{v}})$ were $(40,20)$ and $(45,70)$, respectively.
For both methods, the trajectories in Fig. \ref{fig:5} were generated by the policies using the parameters from the training run that gave the largest number of successful problems.
Fig. \ref{fig:5} shows that the proposed method tended to settle the inter-vehicle distance within the target range earlier than AIRL-BMP and to give trajectories with smaller oscillations, as was the case for the training problems.
Table~\ref{tab:chara-eva} quantifies these characteristics for the successful evaluation problems.
For the first three measures, the values were averaged over the successful problems for each set of learned parameters, and Table~\ref{tab:chara-eva} shows the mean and the standard deviation of these averages over the 20 sets.
The last measure is pooled over all the successful problems in all the sets.
A Mann-Whitney U test between the proposed method and AIRL-BMP showed that the differences in all four measures were significant ($p<0.001$, $n=20$ for each method), where the pooled standard deviation for each set was used for the last measure.
These results suggest that the proposed method acquired higher generalization performance than AIRL-BMP, capable of generating trajectories for successful tasks that settled earlier and with smaller oscillations.

\section{Conclusion}
\label{sec:conc}

This paper proposed a fuzzy control method that integrates a reward estimation mechanism based on the AIRL framework into the base model, in which reinforcement learning is integrated with fuzzy inference.
In the proposed method, the policy of the base model is treated as the generator, and the set of membership functions in the policy is also used as the cost function of the discriminator.
This cost function approximates a quantity corresponding to the advantage of an action with respect to the optimal policy, using the set of membership functions rather than a value function.
With this configuration, the proposed method makes stable and efficient learning possible by performing the discriminator update for reward estimation and the generator update for policy estimation simultaneously on the parameters shared by both, without estimating a value function.

Simulations of an illustrative vehicle speed control task were conducted to compare the performance of the proposed method with that of AIRL-BMP, which uses the discriminator of AIRL as it stands and takes the policy of the base model as the generator.
The results were also compared with those reported on the same problems in previous work for the base model, which uses a reward function set manually in advance.
For the training problems, the proposed method achieved an average task success rate almost equal to that of the base model and much higher than that of AIRL-BMP, with far fewer epochs, and the variation in the success rate across training runs was smaller for the proposed method than for the other methods.
For the evaluation problems, the proposed method achieved the highest average task success rate among the three methods, $100\%$, with no variation across the 20 sets of results.
These results show that the proposed method achieves stable and efficient learning and can acquire policies with high generalization performance.

In this paper, the proposed method was verified on an illustrative vehicle speed control task.
Applying the method to more realistic problems, in which the task and the shapes of the membership functions differ from those in this example, is one of the next steps.
The simultaneous learning of the parameters of the membership functions in the consequent part and of the rule weights was not addressed in this paper, and will be examined in future work.

\backmatter

\bmhead{Funding}
No funding was received to assist with the preparation of this manuscript.

\bmhead{Data availability}
The datasets generated during and/or analysed during the current study are available from the corresponding author on reasonable request.

\section*{Declarations}

\bmhead{Conflict of interest}
The author has no conflicts of interest to declare that are relevant to the content of this article.



\begin{appendices}

\section{Derivation of Eq. (\ref{eq:16})}
\label{app:dev16}
\setcounter{equation}{0}
\renewcommand{\theHequation}{A\arabic{equation}}

Substituting Eq. (\ref{eq:14}) into the right-hand side of Eq. (\ref{eq:15}) gives
\begin{align}
& \mathcal{L}(\alpha^i_j) \!=\!\!\! \sum_{t=0}^{L-1}\!\!\left\{\!-\mathbb{E}_{\mathcal{D}_{t}}\!\!\left[\!f_{\!\alpha^i_j}\!(x,y,x')\right.\right.\nonumber\\
& \quad\quad\quad\quad\quad\quad\quad\ \ \left.-\ln\!\!\left(\!\exp(f_{\!\alpha^i_j}\!(x,y,x')\!)\!+\!\pi(y\!\!\mid\!\! x)\!\right)\!\right]\nonumber\\
& \quad\quad\quad\quad\quad -\!\mathbb{E}_{\pi_t}\!\Big[\ln(\pi(y\!\mid\! x)\!)\nonumber\\
& \quad\quad\quad\quad\quad\quad\quad\ \ \left.\!-\ln\!\!\left(\!\exp(f_{\!\alpha^i_j}\!(x,y,x')\!)\!+\!\pi(y\!\!\mid\!\! x)\!\right)\!\!\Big]\!\right\}\nonumber\\
\label{eq:A1}
\end{align}
Treating $\pi$ as a constant and differentiating Eq. (\ref{eq:A1}) partially with respect to $\alpha^i_j$ gives
\begin{align}
& \frac{\partial}{\partial\alpha^i_j}\mathcal{L}(\alpha^i_j)\nonumber\\
& =\!\!\!\sum_{t=0}^{L-1}\!\!\left\{\!\!-\mathbb{E}_{\mathcal{D}_{t}}\!\!\left[\!\frac{\partial}{\partial\alpha^i_j}f_{\!\alpha^i_j}\!(x,y,x')\right.\right.\nonumber\\
& \quad\quad\quad\quad\quad \left.-\frac{\exp(f_{\!\alpha^i_j}\!(x,y,x'))\frac{\partial}{\partial\alpha^i_j}f_{\!\alpha^i_j}\!(x,y,x')}{\exp(f_{\!\alpha^i_j}\!(x,y,x'))+\pi(y\!\mid\! x)}\right]\nonumber\\
& \quad\quad\ \, \left.+\mathbb{E}_{\pi_t}\!\!\left[\frac{\exp(f_{\!\alpha^i_j}\!(x,y,x'))\frac{\partial}{\partial\alpha^i_j}f_{\!\alpha^i_j}\!(x,y,x')}{\exp(f_{\!\alpha^i_j}\!(x,y,x'))+\pi(y\!\mid\! x)}\right]\!\right\}
\label{eq:A2}\\
& \nonumber\\
& = \!\!\!\sum_{t=0}^{L-1}\!\!\left\{\!\!-\mathbb{E}_{\mathcal{D}_{t}}\!\!\left[\!(1-D^i_j(x,y,x'))\frac{\partial}{\partial\alpha^i_j}f_{\!\alpha^i_j}\!(x,y,x')\!\right]\right.\nonumber\\
& \quad\quad\ \,\left.+\mathbb{E}_{\pi_t}\!\!\left[D^i_j(x,y,x')\frac{\partial}{\partial\alpha^i_j}f_{\alpha^i_j}(x,y,x')\!\right]\!\right\}
\label{eq:A3}
\end{align}
Multiplying the right-hand side of Eq. (\ref{eq:A3}) by $-\delta$ gives Eq. (\ref{eq:16}).

\section{Derivation of Eq. (\ref{eq:21})}
\label{app:dev21}
\setcounter{equation}{0}
\renewcommand{\theHequation}{B\arabic{equation}}

From Eq. (\ref{eq:6}), the characteristic eligibility $e_{\alpha^i_j}(t)$ is expressed as
\begin{align}
& e_{\alpha^i_j}(t)\nonumber\\
& =\! -\frac{1}{T}\!\left\{\!\!\frac{\partial E(y(t);\!x(t)\!)}{\partial \alpha^i_j}-\mathbb{E}_{\pi(y\mid x(t)\!)}\!\!\left[\!\frac{\partial E(y;\!x(t)\!)}{\partial \alpha^i_j}\!\right]\!\right\}
\label{eq:A4}
\end{align}

From Eq. (\ref{eq:2}), the first term in the braces on the right-hand side of Eq. (\ref{eq:A4}) is
\begin{equation}
\frac{\partial E(y(t);x(t))}{\partial \alpha^i_j} = -\eta_i B^i(y(t))\frac{\partial}{\partial \alpha^i_j}A^i(x(t))
\label{eq:A5}
\end{equation}

By replacing $y(t)$ with $y$ in Eq. (\ref{eq:A5}) and taking the expectation of both sides, the second term in the braces on the right-hand side of Eq. (\ref{eq:A4}) is expressed as
\begin{align}
& \mathbb{E}_{\pi(y\mid x(t))}\!\!\left[\!\frac{\partial E(y;\!x(t))}{\partial \alpha^i_j}\!\right]\nonumber\\
& = -\eta_i \mathbb{E}_{\pi(y\mid x(t))}\left[B^i(y)\right]\frac{\partial}{\partial \alpha^i_j}A^i(x(t))
\label{eq:A6}
\end{align}
Substituting Eqs. (\ref{eq:A5}) and (\ref{eq:A6}) into the right-hand side of Eq. (\ref{eq:A4}) gives Eq. (\ref{eq:21}).

\section{Update Equations Used in the Computer Experiments}
\label{app:upd-eqs}
\setcounter{equation}{0}
\renewcommand{\theHequation}{C\arabic{equation}}
\setcounter{table}{0}
\renewcommand{\theHtable}{C\arabic{table}}

\subsection{Discriminator in the Proposed Method}
\label{subapp:disc-pro}

From Eqs. (\ref{eq:12}) and (\ref{eq:13}), the cost function for $(x(t),y(t),x(t+1))$ contained in the trajectory of the demonstration is
\begin{align}
& f_{\alpha^i_j}(x(t),y(t),x(t+1))\nonumber\\
& = \ln\!\left\{1-\left|\mathbb{E}_{\mathcal{D}_{t}}\!\!\left[ A^i_j(x'_j;\alpha^i_j)\right]-A^i_j(x_{j}(t+1);\alpha^i_j) \right| \right\}\nonumber\\
\label{eq:A7}
\end{align}
When the trajectory of the demonstration for each problem consists of one episode, $\mathbb{E}_{\mathcal{D}_{t}}[ A^i_j(x'_j;\alpha^i_j)]=A^i_j(x_{j}(t+1);\alpha^i_j)$ holds, and the right-hand side of Eq. (\ref{eq:A7}) is therefore zero.
The gradient of Eq. (\ref{eq:A7}) with respect to the parameter $\alpha^i_j$ is then also zero, and the first term in the braces on the right-hand side of Eq. (\ref{eq:16}) is zero.
Consequently, in the computer experiments with the learning rule of the proposed method, the update equation for the parameters $\alpha^i_j\in\{\alpha^{\textrm{long}}_1, \alpha^{\textrm{short}}_1, \alpha^{\textrm{fast}}_2, \alpha^{\textrm{slow}}_2\}$ in the cost function $f_{\alpha^i_j}$ of the discriminator $D^i_j$ is
\begin{align}
\Delta\alpha^i_j &= \delta\!\sum_{t=0}^{L-1}\!\left\{\!-\mathbb{E}_{\pi_t}\!\!\left[\!D^i_j(x,y,x')\frac{\partial}{\partial\alpha^i_j}f_{\!\alpha^i_j}(x,y,x')\!\right]\!\right\}\nonumber\\
&= \delta\!\sum_{t=0}^{L-1}\!\left\{\!\mathbb{E}_{\pi_t}\!\!\left[\frac{\frac{\partial}{\partial\alpha^i_j}| v_{\alpha^i_j}(x';x,y)|}{1-| v_{\alpha^i_j}(x';x,y)|+\pi(y\!\mid\! x)}\!\right]\!\right\}\nonumber\\
\label{eq:A8}
\end{align}

In these experiments, $A^{\textrm{long}}_1(\boldsymbol{x}_1)$ and $A^{\textrm{fast}}_2(\boldsymbol{x}_2)$ were assumed to be monotonically increasing, and $A^{\textrm{short}}_1(\boldsymbol{x}_1)$ and $A^{\textrm{slow}}_2(\boldsymbol{x}_2)$ were assumed to be monotonically decreasing.
Under these assumptions, the sign of $v_{\alpha^i_j}$ is determined by the magnitude relation between $x_{\textrm{d}}$ or $x_{\textrm{v}}$ of the demonstration and that of the sample.
The gradient of $| v_{\alpha^i_j}|$ was therefore computed as shown in Table~\ref{tab:A1}.

\begin{table}[t]
\caption{Gradient of $|v_{\alpha^i_j}|$ in each case}\label{tab:A1}
\begin{tabular}{@{}cccc@{}}
\toprule
\multicolumn{2}{c}{demonstration $\geq$ sample}
 & \multicolumn{2}{c}{demonstration $<$ sample} \\
\cmidrule(r){1-2}\cmidrule(l){3-4}
$x_{\textrm{d}}$ & $x_{\textrm{v}}$
 & $x_{\textrm{d}}$ & $x_{\textrm{v}}$ \\
\midrule
\,\ $\dfrac{\partial v_{\alpha^{\textrm{long}}_1}}{\partial\alpha^{\textrm{long}}_1}$
 & \,\ $\dfrac{\partial v_{\alpha^{\textrm{fast}}_2}}{\partial\alpha^{\textrm{fast}}_2}$
 & $-\dfrac{\partial v_{\alpha^{\textrm{long}}_1}}{\partial\alpha^{\textrm{long}}_1}$
 & $-\dfrac{\partial v_{\alpha^{\textrm{fast}}_2}}{\partial\alpha^{\textrm{fast}}_2}$ \\
\addlinespace
$-\dfrac{\partial v_{\alpha^{\textrm{short}}_1}}{\partial\alpha^{\textrm{short}}_1}$
 & $-\dfrac{\partial v_{\alpha^{\textrm{slow}}_2}}{\partial\alpha^{\textrm{slow}}_2}$
 & \,\ $\dfrac{\partial v_{\alpha^{\textrm{short}}_1}}{\partial\alpha^{\textrm{short}}_1}$
 & \,\ $\dfrac{\partial v_{\alpha^{\textrm{slow}}_2}}{\partial\alpha^{\textrm{slow}}_2}$ \\
\botrule
\end{tabular}
\end{table}

\begin{table*}[!t]
\centering
\begin{threeparttable}
\setcounter{table}{0}
\renewcommand{\thetable}{D\arabic{table}}
\renewcommand{\theHtable}{D\arabic{table}}
\caption{Node values defining the consequent membership functions for the accelerator and brake inputs}\label{tab:mf}
\tablebodyfont
\setlength{\tabcolsep}{6pt}
\begin{tabular}{@{}lccccccccccc@{}}
\toprule
$y_1$ & $-5$ & $-4$ & $-3$ & $-2$ & $-1$ & $0$ & $1$ & $2$ & $3$ & $4$ & $5$ \\
\midrule
$B_1^{\textrm{high-ac.}}$ & 0   & 0   & 0   & 0   & 0   & 0 & 0.2 & 0.3 & 0.6 & 0.8 & 0.9 \\
$B_1^{\textrm{low-ac.}}$  & 0   & 0   & 0   & 0   & 0   & 1 & 0.8 & 0.6 & 0.5 & 0.4 & 0.2 \\
$B_1^{\textrm{high-br.}}$ & 0.9 & 0.7 & 0.5 & 0.3 & 0.2 & 0 & 0   & 0   & 0   & 0   & 0   \\
$B_1^{\textrm{low-br.}}$  & 0.1 & 0.2 & 0.4 & 0.6 & 0.7 & 1 & 0   & 0   & 0   & 0   & 0   \\
\botrule
\end{tabular}
\end{threeparttable}
\end{table*}

\subsection{Generator in the Proposed Method}
\label{subapp:gen-pro}

From Eq. (\ref{eq:21}), in the computer experiments with the learning rule of the proposed method, the characteristic eligibility for the policy of the generator is expressed as
\begin{align}
e_{\alpha^i_j}(t) &= -\frac{1}{T}\eta_i\left\{\mathbb{E}_{\pi(y_1\mid x(t))}\left[B^i_1(y_1)\right]-B^i_1(y_1(t))\right\}\nonumber\\
& \quad\cdot\frac{\partial}{\partial \alpha^i_j}\left\{A^i_1(x_1(t))A^i_2(x_2(t))\right\}
\label{eq:A9}
\end{align}
Eqs. (\ref{eq:18}) and (\ref{eq:19}) for the case $\mu=\alpha^i_j$ and Eq. (\ref{eq:A9}) are substituted into Eq. (\ref{eq:20}). 
For $q=0$ and $\gamma=1$, the update equation for the parameters $\alpha^i_j\in\{\alpha^{\textrm{long}}_1, \alpha^{\textrm{short}}_1, \alpha^{\textrm{fast}}_2, \alpha^{\textrm{slow}}_2\}$ in the membership functions $A^i_j$ that constitute the policy $\pi$ of the generator is then expressed as
\begin{align}
& \Delta\alpha^i_j\nonumber\\
&= \!\varepsilon \mathbb{E}_{\tau\sim\pi}\!\Biggl[\sum_{t=0}^{L-1}\!\Biggl\{\!\sum_{u=0}^{L-1}\!\!\left(\ln(D^i_j(x(u),y(u),x(u\!+\!1)))\right.\nonumber\\
& \quad\quad\quad\quad\left.-\ln(1\!-\!D^i_j(x(\!u\!)\!,y(\!u\!)\!,x(\!u\!+\!1\!)\!)\!)\right)\!\!\Biggr\}e_{\alpha^i_j}\!(t)\!\Biggr]\nonumber\\
&= \!\varepsilon \mathbb{E}_{\tau\sim\pi}\!\Biggl[\sum_{t=0}^{L-1}\!\Biggl\{\!\sum_{u=0}^{L-1}\!\bigl(f_{\alpha^i_j}(x(u),y(u),x(u\!+\!1))\nonumber\\
& \quad\quad\quad\quad-\!\ln(\pi(y(u)\!\mid\! x(u)))\bigr)\!\Biggr\}e_{\alpha^i_j}\!(t)\!\Biggr]
\label{eq:A10}
\end{align}

For example, in the case of rule $i=5$, from Eqs. (\ref{eq:23}), (\ref{eq:24}), and (\ref{eq:25}), the characteristic eligibility $e_{\alpha^5_1}(t)=e_{\alpha^{\textrm{long}}_1}(t)$ is
\begin{align}
& e_{\alpha^{\textrm{long}}_1}(t)\nonumber\\
&= -\frac{1}{T}\eta_5 \!\left\{\mathbb{E}_{\pi(y_1\mid x(t))}\!\left[B^{\textrm{no-op.}}_1(y_1)\right]\!-\!B^{\textrm{no-op.}}_1(y_1(t))\right\}\nonumber\\
& \quad\cdot A^{\textrm{fast}}_2(x_2(t))\frac{\partial A^{\textrm{long}}_1(x_1(t))}{\partial \alpha^{\textrm{long}}_1}
\label{eq:A11}
\end{align}
Substituting Eq. (\ref{eq:A11}) into Eq. (\ref{eq:A10}) gives the update equation for the parameter $\alpha^5_1=\alpha^{\textrm{long}}_1$ in the generator.

\subsection{Discriminator in AIRL-BMP}
\label{subapp:disc-airl}

From Eqs. (\ref{eq:8}) and (\ref{eq:10}), the update equations for the parameters $\theta$ and $\phi$ in the cost function $f_{\theta,\phi}$ of the discriminator $D$ in AIRL-BMP are
\begin{flalign}
\Delta\theta = \delta\!\sum_{t=0}^{L-1}\! &\left\{\mathbb{E}_{\mathcal{D}_{t}}\!\!\left[(1-D(x,y,x'))\frac{\partial}{\partial\theta}g(x,y)\right]\right.\nonumber\\
&\ \left.-\mathbb{E}_{\pi_t}\!\!\left[D(x,y,x')\frac{\partial}{\partial\theta}g(x,y)\right]\!\right\}
\label{eq:A12}
\end{flalign}
\begin{align}
\Delta\phi\! =\! \delta\!\!\sum_{t=0}^{L-1}\!\!&\left\{\!\mathbb{E}_{\mathcal{D}_{t}}\!\!\left[\!(1\!-\!D(x,y,x')\!)\frac{\partial}{\partial\phi}(\kappa h(x')\!-\!h(x)\!)\!\right]\right.\nonumber\\
&\ \left.-\mathbb{E}_{\pi_t}\!\!\left[D(x,y,x')\frac{\partial}{\partial\phi}(\kappa h(x')-h(x))\right]\!\right\}
\label{eq:A13}
\end{align}

\subsection{Generator in AIRL-BMP}
\label{subapp:gen-airl}

Eqs. (\ref{eq:11}) and (\ref{eq:A9}) are substituted into Eq. (\ref{eq:20}).
For $q=0$ and $\gamma=1$, the update equation for the parameters $\alpha^i_j\in\{\alpha^{\textrm{long}}_1, \alpha^{\textrm{short}}_1, \alpha^{\textrm{fast}}_2, \alpha^{\textrm{slow}}_2\}$ in the membership functions $A^i_j$ that constitute the policy $\pi$ of the generator in AIRL-BMP is then expressed as
\begin{align}
&\Delta\alpha^i_j\nonumber\\
&= \varepsilon \mathbb{E}_{\tau\sim\pi}\!\Biggl[\sum_{t=0}^{L-1}\!\Biggl\{\!\sum_{u=0}^{L-1}\!\bigl(\ln(D(x(u),y(u),x(u\!+\!1)))\nonumber\\
& \quad\quad\quad\quad-\!\ln(1\!-\!D(x(u),y(u),x(u\!+\!1)))\bigr)\!\!\Biggr\}e_{\!\alpha^i_j}\!(t)\!\Biggr]\nonumber\\
&= \varepsilon \mathbb{E}_{\tau\sim\pi}\!\Biggl[\sum_{t=0}^{L-1}\!\Biggl\{\!\sum_{u=0}^{L-1}\bigl(f_{\theta,\phi}(x(u),y(u),x(u\!+\!1))\nonumber\\
& \quad\quad\quad\quad-\!\ln(\pi(y(u)\!\mid\! x(u)))\bigr)\!\!\Biggr\}e_{\!\alpha^i_j}\!(t)\!\Biggr]
\label{eq:A14}
\end{align}

\section{Definition of the Consequent Membership Functions}
\label{app:mfc}
\setcounter{table}{0}
\renewcommand{\theHtable}{D\arabic{table}}

The output variable $y_1$ was discretized with a step of 0.1 over $[-5, 5]$. The membership functions for the accelerator and brake inputs were defined as the piecewise linear interpolation of the node values listed in Table~\ref{tab:mf}, with the functions for the accelerator input zero for $y_1<0$ and those for the brake input zero for $y_1>0$. $B_1^{\textrm{no-op.}}$ takes the value 1 only at $y_1=0$.

\end{appendices}

\bibliography{references}

\end{document}